\documentclass[10pt,twocolumn,letterpaper]{article}

\usepackage[pagenumbers]{cvpr} 

\usepackage{graphicx}
\usepackage{amsmath}
\usepackage{amssymb}
\usepackage{booktabs}

\usepackage{hyperref}
\hypersetup{pagebackref,breaklinks,colorlinks}

\usepackage[capitalize]{cleveref}
\crefname{section}{Sec.}{Secs.}
\Crefname{section}{Section}{Sections}
\Crefname{table}{Table}{Tables}
\crefname{table}{Tab.}{Tabs.}

\usepackage[dvipsnames]{xcolor}
\usepackage{enumitem}
\usepackage{multirow}
\usepackage{pifont}
\usepackage{soul}
\usepackage[accsupp]{axessibility}
\usepackage[ruled,vlined,noend,linesnumbered]{algorithm2e}
\usepackage{caption}
\usepackage{subcaption}

\let\svthefootnote\thefootnote
\newcommand\freefootnote[1]{%
  \let\thefootnote\relax%
  \footnotetext{#1}%
  \let\thefootnote\svthefootnote%
}
\SetKwInput{KwInput}{Input}                
\SetKwInput{KwOutput}{Output}              

\def\cvprPaperID{9938} 
\def\confName{CVPR}
\def\confYear{2023}

\begin{document}

\title{Contrastive Mean Teacher for Domain Adaptive Object Detectors}

\author{Shengcao Cao\textsuperscript{\textrm 1}\quad Dhiraj Joshi\textsuperscript{\textrm 2}\quad Liang-Yan Gui\textsuperscript{\textrm 1}\quad Yu-Xiong Wang\textsuperscript{\textrm 1}\\
\textsuperscript{\textrm 1}University of Illinois at Urbana-Champaign\quad\textsuperscript{\textrm 2}IBM Research\\
{\tt\small\textsuperscript{\textrm 1}\{cao44,lgui,yxw\}@illinois.edu\quad\textsuperscript{\textrm 2}djoshi@us.ibm.com}}

\maketitle

\begin{abstract}
Object detectors often suffer from the domain gap between training (source domain) and real-world applications (target domain). Mean-teacher self-training is a powerful paradigm in unsupervised domain adaptation for object detection, but it struggles with low-quality pseudo-labels. In this work, we identify the intriguing alignment and synergy between mean-teacher self-training and contrastive learning. Motivated by this, we propose Contrastive Mean Teacher (CMT) -- a unified, general-purpose framework with the two paradigms naturally integrated to maximize beneficial learning signals. Instead of using pseudo-labels solely for final predictions, our strategy extracts object-level features using pseudo-labels and optimizes them via contrastive learning, without requiring labels in the target domain. When combined with recent mean-teacher self-training methods, CMT leads to new state-of-the-art target-domain performance: 51.9\% mAP on Foggy Cityscapes, outperforming the previously best by 2.1\% mAP. Notably, CMT can stabilize performance and provide more significant gains as pseudo-label noise increases.
\end{abstract}

\freefootnote{\scriptsize Code available at \url{https://github.com/Shengcao-Cao/CMT}}

\section{Introduction}
\label{sec:intro}
The domain gap between curated datasets (source domain) and real-world applications (target domain, \eg, on edge devices or robotic systems) often leads to deteriorated performance for object detectors. Meanwhile, accurate labels provided by humans are costly or even unavailable in practice. Aiming at maximizing performance in the target domain while minimizing human supervision, unsupervised domain adaptation mitigates the domain gap via adversarial training~\cite{chen2018domain,saito2019strong}, domain randomization~\cite{kim2019diversify}, image translation~\cite{inoue2018cross,hsu2020progressive,chen2020harmonizing}, \etc.

\begin{figure}
    \centering
    \includegraphics[width=\columnwidth]{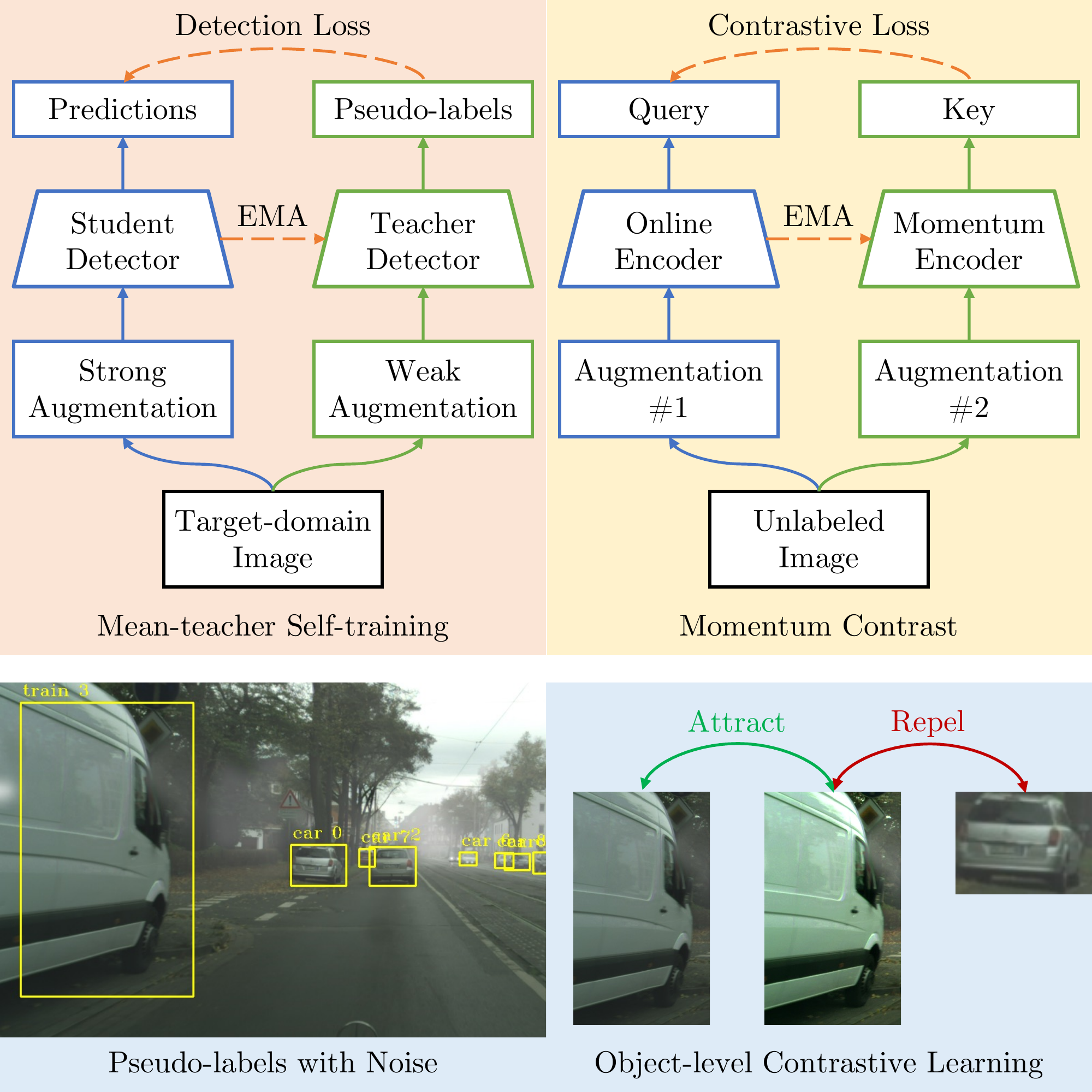}
    \caption{\textbf{Overview of Contrastive Mean Teacher.}
    \textbf{Top:} Mean-teacher self-training~\cite{cai2019exploring,deng2021unbiased,li2022cross,chen2022learning} for unsupervised domain adaptation (left) and Momentum Contrast~\cite{he2020momentum} for unsupervised representation learning (right) share the same underlying structure, and thus can be naturally integrated into our unified framework, \emph{Contrastive Mean Teacher}.
    \textbf{Bottom:} Contrastive Mean Teacher benefits unsupervised domain adaptation even when pseudo-labels are noisy. In this example, the teacher detector incorrectly detects the truck as a train and the bounding box is slightly off. Reinforcing this wrong pseudo-label in the student harms the performance. Contrarily, our proposed {\em object-level} contrastive learning still finds meaningful learning signals from it, by enforcing feature-level similarities between the same objects and dis-similarities between different ones.}
    \label{fig:teaser}
    \vspace{-3mm}
\end{figure}

In contrast to the aforementioned techniques that explicitly model the domain gap, state-of-the-art domain adaptive object detectors~\cite{li2022cross,chen2022learning} follow a mean-teacher self-training paradigm~\cite{cai2019exploring,deng2021unbiased}, which explores a teacher-student mutual learning strategy to gradually adapt the object detector for cross-domain detection. As illustrated in Figure~\ref{fig:teaser}-top, the teacher generates pseudo-labels from detected objects in the target domain, and the pseudo-labels are then used to supervise the student's predictions. In return, the teacher's weights are updated as the exponential moving average (EMA) of the student's weights.

Outside of unsupervised domain adaptation, contrastive learning~\cite{chen2020simple,he2020momentum,grill2020bootstrap,caron2020unsupervised} has served as an effective approach to learning from unlabeled data. Contrastive learning optimizes feature representations based on the similarities between instances in a fully self-supervised manner. {\em Intriguingly}, as shown in Figure~\ref{fig:teaser}-top, there in fact exist strong {\em alignment and synergy} between the Momentum Contrast paradigm~\cite{he2020momentum} from contrastive learning and the mean-teacher self-training paradigm~\cite{cai2019exploring,deng2021unbiased} from unsupervised domain adaptation: The momentum encoder (teacher detector) provides stable learning targets for the online encoder (student detector), and in return the former is smoothly updated by the latter's EMA. Inspired by this observation, we propose \emph{Contrastive Mean Teacher (CMT)} -- a unified framework with the two paradigms naturally integrated. We find that their benefits can compound, especially with contrastive learning facilitating the feature adaptation towards the target domain from the following aspects.

First, mean-teacher self-training suffers from the poor quality of pseudo-labels, but contrastive learning does not rely on accurate labels. Figure~\ref{fig:teaser}-bottom shows an illustrative example: On the one hand, the teacher detector produces pseudo-labels in the mean-teacher self-training framework, but they can never be perfect (otherwise, domain adaptation would not be needed). The student is trained to fit its detection results towards these noisy pseudo-labels. Consequently, mis-predictions in the pseudo-labels become harmful learning signals and limit the target-domain student performance. On the other hand, contrastive learning does not require accurate labels for learning. Either separating individual instances~\cite{chen2020simple,he2020momentum} or separating instance clusters~\cite{caron2020unsupervised} (which do not necessarily coincide with the actual classes) can produce powerful representations. Therefore, CMT effectively learns to adapt its features in the target domain, even with noisy pseudo-labels.

Second, by introducing an {\em object-level} contrastive learning strategy, we learn more fine-grained, localized representations that are crucial for object detection. Traditionally, contrastive learning treats data samples as monolithic instances but ignores the complex composition of objects in natural scenes. This is problematic as a natural image consists of multiple heterogeneous objects, so learning one homogeneous feature may not suffice for object detection. Hence, some recent contrastive learning approaches learn representations at the pixel~\cite{van2021unsupervised}, region~\cite{bai2022point}, or object~\cite{wei2021aligning} levels, for object detection {\em yet without considering the challenging scenario of domain adaptation}. Different from such prior work, in CMT we propose object-level contrastive learning to precisely adapt localized features to the target domain. In addition, we exploit predicted classes from noisy pseudo-labels, and further augment our object-level contrastive learning with {\em multi-scale} features, to maximize the beneficial learning signals.

Third, CMT is a general-purpose framework and can be readily combined with existing work in mean-teacher self-training. The object-level contrastive loss acts as a {\em drop-in enhancement} for feature learning, and does not change the original training pipelines. Combined with the most recent methods (\eg, Adaptive Teacher~\cite{li2022cross}, Probabilistic Teacher~\cite{chen2022learning}), we achieve new state-of-the-art performance in unsupervised domain adaptation for object detection.

To conclude, our contributions include:
\begin{itemize}[leftmargin=*, noitemsep, nolistsep]
    \item We identify the intrinsic alignment and synergy between contrastive learning and mean-teacher self-training, and propose an integrated unsupervised domain adaptation framework, Contrastive Mean Teacher (CMT).
    \item We develop a general-purpose object-level contrastive learning strategy to enhance the representation learning in unsupervised domain adaptation for object detection. Notably, the benefit of our strategy becomes more pronounced with increased pseudo-label noise (see Figure~\ref{fig:noise}).
    \item We show that our proposed framework can be combined with several existing mean-teacher self-training methods without effort, and the combination achieves state-of-the-art performance on multiple benchmarks, \eg, improving the adaptation performance on Cityscapes to Foggy Cityscapes from 49.8\% mAP to 51.9\% mAP.
\end{itemize}

\section{Related Work}
\label{sec:related}
\noindent\textbf{Unsupervised domain adaptation for object detection.}
Unsupervised domain adaptation is initially studied for image classification~\cite{ganin2016domain}, and recently extended to object detection applications.
Adversarial feature learning methods~\cite{chen2018domain,saito2019strong,xu2020exploring,vs2021mega} employ a domain discriminator and train the feature encoder and discriminator adversarially, so that domain-invariant visual features can be learned.
Image-to-image translation methods~\cite{inoue2018cross,hsu2020progressive,chen2020harmonizing} synthesize source-like images from target-domain contents (or the other way around) using generative models (\eg, CycleGAN~\cite{zhu2017unpaired}) to mitigate domain gaps.
More recently, the idea of Mean Teacher~\cite{tarvainen2017mean} is extended from semi-supervised object detection to unsupervised domain adaptation for object detection by \cite{cai2019exploring}. Following this exploration, Unbiased Mean Teacher (UMT)~\cite{deng2021unbiased} integrates image translation with Mean Teacher, Adaptive Teacher~\cite{li2022cross} applies weak-strong augmentation and adversarial training, and Probabilistic Teacher (PT)~\cite{chen2022learning} improves pseudo-labeling with uncertainty-guided self-training for both classification and localization. Though this line of research plays a leading role in unsupervised domain adaptation for object detection, the major challenge still comes from the poor quality of pseudo-labels generated by Mean Teacher.
For a comprehensive overview of this topic, one may refer to \cite{oza2021unsupervised}.

\noindent\textbf{Contrastive learning.}
Contrastive loss~\cite{hadsell2006dimensionality} measures the representation similarities between sample pairs. Recently, contrastive learning successfully powers self-supervised visual representation pre-training, with the help of a large batch size~\cite{chen2020simple}, memory bank~\cite{he2020momentum}, asymmetric architecture~\cite{grill2020bootstrap}, or clustering~\cite{caron2020unsupervised}. Self-supervised contrastive learning has outperformed supervised pre-training in some settings~\cite{tomasev2022pushing}.
To align contrastive pre-training with down-stream tasks other than image classification (\eg, object detection, semantic segmentation), more fine-grained approaches have been proposed based on masks~\cite{van2021unsupervised,henaff2021efficient}, objects~\cite{wei2021aligning}, or regions~\cite{bai2022point}.
Our object-level contrastive learning strategy is inspired by this line of research. Instead of applying contrastive learning in pre-training visual backbones, we study how to improve domain adaptive object detectors using noisy pseudo-labels and object-level contrast. Technically, we construct contrastive pairs using the predicted classes in pseudo-labels and optimize multi-scale features, both of which are different from typical object-level contrastive learning. Recently, contrastive learning is explored in teacher-student learning for detection~\cite{yao2021g, vs2023towards}. However, our work is the \emph{first} to analyze the synergy between Mean Teacher~\cite{tarvainen2017mean} and contrastive learning. Moreover, we present a \emph{simple and general} framework CMT that does not rely on negative sample mining or selection.

\section{Approach}
\label{sec:approach}
\begin{figure*}[t]
    \centering
    \includegraphics[width=2\columnwidth]{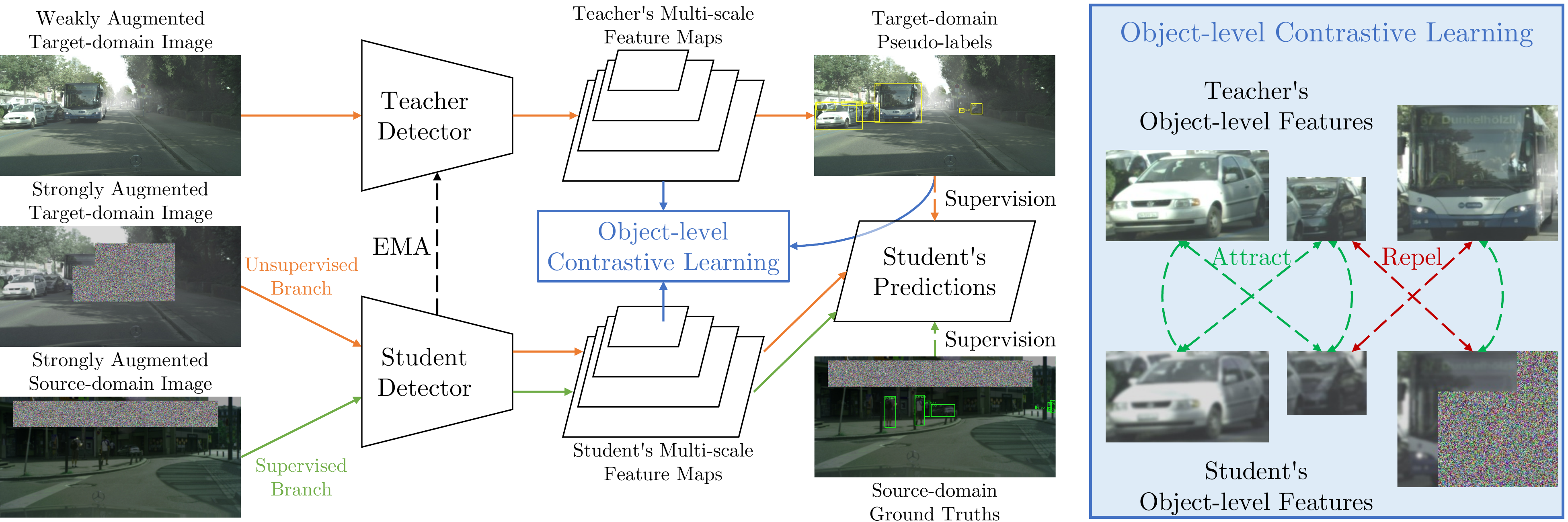}
    \caption{
    \textbf{Our proposed Contrastive Mean Teacher (CMT) framework.}
    \textbf{Left:} Mean-teacher self-training paradigm in unsupervised domain adaptation for object detection. The \textcolor{Orange}{unsupervised branch} uses unlabeled target-domain images and pseudo-labels generated by the teacher, which is updated by the student's exponential moving average (EMA), and performs \textcolor{NavyBlue}{object-level contrastive learning}; the \textcolor{LimeGreen}{supervised branch} uses labeled source-domain images.
    \textbf{Right:} \textcolor{NavyBlue}{Object-level contrastive learning} strategy. Object-level features can be extracted from the teacher's and student's feature maps using the pseudo-labels. Contrastive loss is enforced for refined feature adaptation.}
    \label{fig:main}
    \vspace{-3mm}
\end{figure*}

We introduce our proposed Contrastive Mean Teacher (CMT) in the following steps. In Section~\ref{sec:mean-teacher}, we first describe the mean-teacher self-training paradigm that is shared by recent methods~\cite{cai2019exploring,deng2021unbiased,li2022cross,chen2022learning} in unsupervised domain adaptation for object detection. Then in Section~\ref{sec:moco}, we connect mean-teacher self-training with Momentum Contrast~\cite{he2020momentum}, a typical contrastive learning method, to unify them into one framework, Contrastive Mean Teacher (see Figure~\ref{fig:main}-left). Finally in Section~\ref{sec:obj-contrast}, we introduce the object-level contrastive learning strategy used in CMT (see Figure~\ref{fig:main}-right). We include the pseudo-code for CMT in the supplementary material.

\subsection{Mean-teacher Self-training}
\label{sec:mean-teacher}
We build our approach upon recent unsupervised domain adaptation methods of the mean-teacher self-training paradigm. In this section, we summarize the mutual-learning process in this paradigm.

\noindent\textbf{Overall structure.} This paradigm mainly consists of two detector models of the identical architecture, the teacher and the student. There is mutual knowledge transfer between the two, but the two directions of knowledge transfer are in different forms. Both models take inputs from the target domain. Figure~\ref{fig:teaser}-top-left shows a brief sketch of this mean-teacher self-training paradigm.

\noindent\textbf{Teacher $\rightarrow$ Student knowledge transfer.} The teacher first detects objects in the target-domain input images. Then, pseudo-labels can be generated from the detection results by some post-processing (\eg, filtering by confidence scores and non-maximum suppression). The teacher's knowledge is transferred by fitting the student's predictions towards these pseudo-labels in the target domain. Standard bounding box regression loss and classification loss are minimized in this knowledge transfer. To ensure high quality of the pseudo-labels, the teacher's inputs are weakly augmented (\eg, simple cropping and flipping)~\cite{li2022cross,chen2022learning} or translated to the source-domain style~\cite{deng2021unbiased}. Meanwhile, the student's inputs are strongly augmented (\eg, blurring and color jittering) or not translated to the source-domain style.

\noindent\textbf{Student $\rightarrow$ Teacher knowledge transfer.} The student is updated by minimizing the detection loss with gradient descent. We do not compute gradients for the teacher, though. The teacher's weights $\theta^\mathcal{T}$ are updated as the exponential moving average (EMA) of the student's weights $\theta^\mathcal{S}$:
\begin{equation}
\theta^\mathcal{T}\gets\alpha\theta^\mathcal{T} + (1-\alpha)\theta^\mathcal{S},
\end{equation}
where $\alpha\in[0, 1)$ is a momentum coefficient and is usually large (0.9996 in our setting) to ensure smooth teacher updates.
Therefore, the teacher can be considered as an ensemble of historical students and provides more stable learning targets. The teacher is also used as the model for evaluation, due to its reliable target-domain performance.

\subsection{Aligning Mean-teacher Self-training with Momentum Contrast}
\label{sec:moco}
In this section, we first briefly introduce Momentum Contrast (MoCo)~\cite{he2020momentum}, and then describe the alignment between mean-teacher self-training and MoCo.

\noindent\textbf{Momentum Contrast.} MoCo is a widely used contrastive learning method for unsupervised visual representation learning. Figure~\ref{fig:teaser}-top-right shows the overall pipeline of this method. It has an online encoder $f(\cdot;\theta^\mathcal{Q})$ and a momentum encoder $f(\cdot;\theta^\mathcal{K})$ that share the same architecture but have different weights. Each input image $I_i$ is augmented into two different views $t^\mathcal{Q}(I_i)$ and $t^\mathcal{K}(I_i)$, and then fed into the two encoders to produce features $z^\mathcal{Q}_i=\text{Normalize}\left(f(t^\mathcal{Q}(I_i);\theta^\mathcal{Q})\right)$ and $z^\mathcal{K}_i=\text{Normalize}\left(f(t^\mathcal{K}(I_i);\theta^\mathcal{K})\right)$. The online encoder is optimized by minimizing the contrastive loss:
\begin{equation}
\mathcal{L}_\text{MoCo}=-\log\frac{\exp(z_i^\mathcal{Q}\cdot z_i^\mathcal{K}/\tau)}{\sum_{j\in\mathcal{D}}\exp(z_i^\mathcal{Q}\cdot z_j^\mathcal{K}/\tau)},
\label{eq:moco}
\end{equation}
where $\tau>0$ is a temperature hyper-parameter and $\mathcal{D}$ is a memory bank of other image features. The feature pair $\langle z_i^\mathcal{Q}, z_i^\mathcal{K}\rangle$ in the numerator corresponds to the same original image, so it is called a positive pair; $\langle z_i^\mathcal{Q}, z_j^\mathcal{K}\rangle$ is a negative pair. In MoCo the memory bank contains a large amount of features generated by the momentum encoder in previous iterations, but in this work we find only using features within one image batch is adequate for our task.

The weights of the momentum encoder $\theta^\mathcal{K}$ is updated as the EMA of the online encoder's weights $\theta^\mathcal{Q}$:
\begin{equation}
\theta^\mathcal{K}\gets\alpha\theta^\mathcal{K} + (1-\alpha)\theta^\mathcal{Q}.
\end{equation}

\noindent\textbf{Alignment between two paradigms.} The mean-teacher self-training and MoCo share the same intrinsic structure, though their designated tasks are different (see Figure~\ref{fig:teaser}):
\begin{itemize}[leftmargin=*, noitemsep, nolistsep]
    \item Two networks of the same architecture are learned jointly. The teacher detector (momentum encoder) is updated as the EMA of the student detector (online encoder), while the latter is updated by gradient descent of minimizing the detection loss (contrastive loss).
    \item The image needs no label. It is augmented differently by $t^\mathcal{S},t^\mathcal{T}$ ($t^\mathcal{Q},t^\mathcal{K}$) into different views. However, the object classes and locations (semantic information) stay the same in two views, so that supervision can be enforced.
    \item The teacher detector (momentum encoder) produces stable learning targets, because it evolves smoothly due to a large $\alpha$ and can be considered as an ensemble of previous models. In mean-teacher self-training, the teacher's data augmentation encourages stable pseudo-labels as well.
\end{itemize}

Therefore, we can naturally integrate the two paradigms into one unified framework, \emph{Contrastive Mean Teacher} (CMT, as shown in Figure~\ref{fig:main}). Since our focused task is still unsupervised domain adaptation for object detection, the main body of CMT follows the mean-teacher self-training paradigm as described in Section~\ref{sec:mean-teacher}, and contrastive learning is combined into it as a drop-in enhancement for feature adaptation. Specifically, we introduce an object-level contrastive learning strategy in CMT.

\subsection{Object-level Contrastive Learning}
\label{sec:obj-contrast}
As described in Section~\ref{sec:mean-teacher}, the teacher generates pseudo-labels from target-domain images for the student to learn. In addition to the supervision at the final prediction level, we make better use of the pseudo-labels to refine the features, via object-level contrastive learning.

\noindent\textbf{Extracting object-level features.} Both the teacher and student take the same image batch $\mathcal{I}$ from the target domain, but may transform $\mathcal{I}$ differently as $t^\mathcal{T}(\mathcal{I})$ and $t^\mathcal{S}(\mathcal{I})$. The teacher generates a set of $N$ pseudo-labels for $\mathcal{I}$, including bounding boxes $\mathcal{B}=\{B_1,\dots,B_N\}$ and predicted classes $\mathcal{C}=\{C_1,\dots,C_N\}$. From the input $t^\mathcal{T}(\mathcal{I})$, we can extract an intermediate feature map $F^\mathcal{T}$ from the teacher's backbone, and similarly get the student's feature map $F^\mathcal{S}$. We use RoIAlign~\cite{he2017mask}, a pooling operation for Regions of Interest (RoI), to extract object-level features and normalize them following the common practice~\cite{chen2020simple,he2020momentum}: $z^\mathcal{M}_i=\text{Normalize}(\text{ROIAlign}(F^\mathcal{M},B_i))$, where the model $\mathcal{M}\in\{\mathcal{T},\mathcal{S}\}$. If $t^\mathcal{T}$ and $t^\mathcal{S}$ change bounding boxes differently, we need to transform $B_i$ to align two feature maps.

\noindent\textbf{Class-based contrast.} We perform contrastive learning between the teacher's and student's object-level features. Inspired by supervised contrastive learning~\cite{khosla2020supervised}, we utilize the teacher's predicted classes to exploit learning signals from pseudo-labels. The contrastive loss is formulated as:
\begin{equation}
\resizebox{\columnwidth}{!}{$
\mathcal{L}_\text{contrast}=\frac{\lambda_\text{contrast}}{N}\sum_{i=1}^N\frac{-1}{|\mathcal{P}(i)|}\sum_{p\in \mathcal{P}(i)}\log\frac{\exp(z_i^\mathcal{S}\cdot z_p^\mathcal{T}/\tau)}{\sum_{j=1}^N\exp(z_i^\mathcal{S}\cdot z_j^\mathcal{T}/\tau)},
$}
\label{eq:contrast}
\end{equation}
where the positive pair set $\mathcal{P}(i)=\{p\mid C_p=C_i, p\in\{1,\dots,N\}\}$ includes all objects of the same predicted class as object $i$, and the balancing weight $\lambda_\text{contrast}>0$ and temperature $\tau>0$ are hyper-parameters. $\mathcal{L}_\text{contrast}$ is added to all other losses (\eg, supervised source-domain detection loss, unsupervised target-domain detection loss) in the mean-teacher self-training method.

\noindent\textbf{Multi-scale features.}
To provide additional learning signals, we perform our object-level contrastive learning at multiple feature levels of the backbone (\eg, VGG~\cite{simonyan2014very}, ResNet~\cite{he2016deep}). For example for the teacher, we have $k$ feature maps $\{F^\mathcal{T}_1,\dots,F^\mathcal{T}_k\}$ of multiple scales. We then scale bounding boxes accordingly, so that they still correspond to the same objects, and extract multi-scale features. The object-level contrastive losses (Equation~\ref{eq:contrast}) at multiple levels are added up and optimized together.

\section{Experiments}
\label{sec:experiment}
\subsection{Datasets and Evaluation}
Our proposed approach Contrastive Mean Teacher (CMT) is evaluated on the following datasets: Cityscapes, Foggy Cityscapes, KITTI, Pascal VOC, and Clipart1k.

\noindent\textbf{Cityscapes}~\cite{cordts2016cityscapes} is a dataset of street scenes. It contains 2,975 training images and 500 validation images, collected from 50 cities. For object detection, we use 8 categories and the bounding boxes are converted from segmentation masks.

\noindent\textbf{Foggy Cityscapes}~\cite{sakaridis2018semantic} is a dataset synthesized from Cityscapes by adding fog to the original images. Three fog levels (0.02, 0.01, 0.005) are simulated corresponding to different visibility ranges. We use the most challenging 0.02 split as well as all splits in our experiments.

\noindent\textbf{KITTI}~\cite{geiger2012we} is another dataset of street scenes, but the data are collected using cameras and in cities that are different from Cityscapes. We use the training split of 7,481 images for domain adaptation, and only consider the car category shared by both KITTI and Cityscapes.

\noindent\textbf{Pascal VOC}~\cite{everingham2010pascal} is a dataset of 20 categories of common objects in realistic scenes. We use the training split of Pascal VOC 2012 containing 11,540 images.

\noindent\textbf{Clipart1k}~\cite{inoue2018cross} is a dataset of clip art images. It shares the same categories as Pascal VOC, but the image style differs. The training and validation splits both have 500 images.

Following prior work, we conduct experiments on three domain adaptation tasks: From normal weather to adverse weather (Cityscapes $\rightarrow$ Foggy Cityscapes), across different cameras (KITTI $\rightarrow$ Cityscapes), and from realistic images to artistic images (Pascal VOC $\rightarrow$ Clipart1k). We use the training splits of both the source domain and the target domain in the unsupervised domain adaptation procedure, and use the validation split of the target domain for performance evaluation. For comparison, we use the mean average precision (mAP) metric with the 0.5 threshold for Intersection over Union (IoU), following the standard practice on the Pascal VOC object detection benchmark.

We consider two base methods in mean-teacher self-training: Adaptive Teacher (AT)~\cite{li2022cross} and Probabilistic Teacher (PT)~\cite{chen2022learning}, since they are state-of-the-art unsupervised domain adaptation methods for object detection. We combine them with our Contrastive Mean Teacher (CMT) framework by adding the object-level contrastive learning objectives to their original adaptation pipelines.

\begin{table*}[t]
    \centering
    \vspace{-3mm}
    \begin{tabular}{@{}l|l|c|c c c c c c c c|c@{}}
        \toprule
        Type & Method & Split & person & rider & car & truck & bus & train & motor & bike & mAP \\
        \midrule
        - & Source$^\dag$ & 0.02 & 22.4 & 26.6 & 28.5 & 9.0 & 16.0 & 4.3 & 15.2 & 25.3 & 18.4 \\
        - & Oracle$^\dag$ & 0.02 & 39.5 & 47.3 & 59.1 & 33.1 & 47.3 & 42.9 & 38.1 & 40.8 & 43.5 \\
        \midrule
        DR & DM~\cite{kim2019diversify} & 0.02 & 30.8 & 40.5 & 40.5 & 27.2 & 38.4 & 34.5 & 28.4 & 32.3 & 34.6 \\
        AFL + IT & HTCN~\cite{chen2020harmonizing} & 0.02 & 33.2 & 47.5 & 47.9 & 31.6 & 47.4 & 40.9 & 32.3 & 37.1 & 39.8 \\
        AFL & MeGA-CDA~\cite{vs2021mega} & 0.02 & 37.7 & 49.0 & 52.4 & 25.4 & 49.2 & 46.9 & 34.5 & 39.0 & 41.8 \\
        AFL & TIA~\cite{zhao2022task} & 0.02 & 34.8 & 46.3 & 49.7 & 31.1 & 52.1 & 48.6 & 37.7 & 38.1 & 42.3 \\
        GR & SIGMA~\cite{li2022sigma} & 0.02 & 46.9 & 48.4 & 63.7 & 27.1 & 50.7 & 35.9 & 34.7 & 41.4 & 43.5 \\
        \midrule
        MT + GR & MTOR~\cite{cai2019exploring} & 0.02 & 30.6 & 41.4 & 44.0 & 21.9 & 43.4 & 40.2 & 31.7 & 33.2 & 35.1 \\
        MT + IT & UMT~\cite{deng2021unbiased} & 0.02 & 33.0 & 46.7 & 48.6 & 34.1 & 56.5 & 46.8 & 30.4 & 37.3 & 41.7 \\
        MT & PT~\cite{chen2022learning} & 0.02 & 40.2 & 48.8 & 59.7 & 30.7 & 51.8 & 30.6 & 35.4 & 44.5 & 42.7 \\
        MT & PT~\cite{chen2022learning} + CMT (Ours) & 0.02 & 42.3 & 51.7 & 64.0 & 26.0 & 42.7 & 37.1 & 42.5 & 44.0 & 43.8 (+1.1) \\
        MT + AFL & AT$^\ddag$~\cite{li2022cross} & 0.02 & 45.3 & 55.7 & 63.6 & 36.8 & 64.9 & 34.9 & 42.1 & 51.3 & 49.3 \\
        MT + AFL & AT~\cite{li2022cross} + CMT (Ours) & 0.02 & 45.9 & 55.7 & 63.7 & 39.6 & 66.0 & 38.8 & 41.4 & 51.2 & \bf{50.3 (+1.0)} \\
        \midrule\midrule
        - & Source$^\dag$ & All & 27.9 & 33.4 & 40.4 & 12.1 & 23.2 & 10.1 & 20.7 & 30.9 & 24.8 \\
        - & Oracle$^\dag$ & All & 41.2 & 49.1 & 61.6 & 32.6 & 56.6 & 49.0 & 37.9 & 42.4 & 46.3\\
        \midrule
        AFL + IT & PDA~\cite{hsu2020progressive} & All & 36.0 & 45.5 & 54.4 & 24.3 & 44.1 & 25.8 & 29.1 & 35.9 & 36.9 \\
        AFL & ICR-CCR~\cite{xu2020exploring} & All & 32.9 & 43.8 & 49.2 & 27.2 & 36.4 & 36.4 & 30.3 & 34.6 & 37.4 \\
        \midrule
        MT & PT~\cite{chen2022learning} & All & 43.2 & 52.4 & 63.4 & 33.4 & 56.6 & 37.8 & 41.3 & 48.7 & 47.1 \\
        MT & PT~\cite{chen2022learning} + CMT (Ours) & All & 45.6 & 55.1 & 66.5 & 34.0 & 59.4 & 42.4 & 43.9 & 47.4 & 49.3 (+2.2) \\
        MT + AFL & AT$^\ddag$~\cite{li2022cross} & All & 46.3 & 55.9 & 64.3 & 38.5 & 61.1 & 39.3 & 40.8 & 52.3 & 49.8 \\
        MT + AFL & AT~\cite{li2022cross} + CMT (Ours) & All & 47.0 & 55.7 & 64.5 & 39.4 & 63.2 & 51.9 & 40.3 & 53.1 & \bf{51.9 (+2.1)} \\
        \bottomrule
    \end{tabular}\\
    \footnotesize{$^\dag$ Results from PT~\cite{chen2022learning}. $^\ddag$ Results reproduced using the released code by AT~\cite{li2022cross} to acquire complete results on the ``0.02'' split.}
    \caption{\textbf{Domain adaptation from normal weather (Cityscapes) to adverse weather (Foggy Cityscapes).} Mean-teacher self-training (``MT'') methods are leading in unsupervised domain adaptation for object detection, outperforming adversarial feature learning (``AFL''), image-to-image translation (``IT''), domain randomization (``DR''), and graph reasoning (``GR'') methods. Our proposed Contrastive Mean Teacher (CMT) consistently improves mean-teacher methods including PT~\cite{chen2022learning} and AT~\cite{li2022cross} on both splits of Foggy Cityscapes (``0.02'' and ``All''), and achieves a new state-of-the-art result of \textbf{51.9\% mAP}. The performance gain of CMT is more significant when more unlabeled training data are available, revealing its potential in improving real-world applications.}
    \label{tab:foggy}
    \vspace{-3mm}
\end{table*}

\subsection{Implementation Details}
For a fair comparison with previous methods, we use the standard Faster R-CNN object detector~\cite{ren2015faster} with the VGG-16~\cite{simonyan2014very} (on Cityscapes) or ResNet-101~\cite{he2016deep} (on Pascal VOC) backbone as the detection model. As for hyper-parameters in all experiments, we set the temperature $\tau=0.07$ (following \cite{he2020momentum,khosla2020supervised}) and balancing weight $\lambda_\text{contrast}=0.05$ (around which we observe only minor performance variations). We extract multi-scale features from the last 4 stages of the backbone networks. Other hyper-parameters are the same as in the original implementation of AT and PT. Our implementation is based on Detectron2~\cite{wu2019detectron2} and the publicly available code by AT and PT. Each experiment is conducted on 4 NVIDIA A100 GPUs.

\noindent\textbf{Post-processing pseudo-labels.} For AT, we observe that some objects are completely erased in the student's view due to the strong augmentation of Cutout~\cite{devries2017improved,zhong2020random}. In such cases, it is no longer meaningful to enforce their features to be similar to those in the teacher's view. We exclude such objects by an empirical criterion: In each object bounding box, we count the pixels where the RGB value difference between the teacher's and student's view is larger than 40. If the ratio of such pixels is higher than 50\%, then the object is considered as removed by Cutout and not included in our object-level contrastive learning. This criterion excludes about one third of all objects. For PT, since the uncertainty-aware pseudo-labels are represented as categorical distributions (for classification) and normal distributions (for location), we need to post-process them to acquire one-hot class labels and bounding boxes for our object-level contrast. We simply take the argmax of the categorical distribution and only keep labels that are foreground and have a confidence score higher than 60\%. The bounding box is constructed from the mean (most possible) of the normal distributions.

\subsection{Adverse Weather}
Object detectors deployed in real-world applications often face a weather condition that is different from the training. For example, the quality of input images captured by cameras may deteriorate when there is rain, snow, or fog. Such adverse weather conditions can be a great challenge to the performance of object detectors. Therefore, we apply domain adaptation methods to overcome this domain shift from normal weather to adverse weather. In this experiment, we evaluate CMT on the commonly used benchmark Cityscapes $\rightarrow$ Foggy Cityscapes, where the object detector needs to adapt from a normal weather condition to a foggy scene with limited visibility.

The results are summarized in Table~\ref{tab:foggy}. To ensure a fair comparison, we provide the results for training and evaluating on both the foggiest images (``0.02'' split) and all synthetic images (``All'' split) in Foggy Cityscapes. As discussed in Section~\ref{sec:related}, the mean-teacher self-training methods~\cite{chen2022learning,li2022cross} are leading unsupervised domain adaptation for object detection. They not only outperform previous non-mean-teacher methods, but also surpass the ``Oracle'' models, which are directly trained in the target domain using ground-truth labels that are not available to unsupervised domain adaptation methods. The reason is that they can leverage the images in both the source domain and the target domain, and transfer cross-domain knowledge.

By combining state-of-the-art mean-teacher methods with our proposed framework CMT, we acquire further performance gain and achieve the best results so far. On both dataset splits of ``0.02'' and ``All,'' we consistently improve two methods PT~\cite{chen2022learning} and AT~\cite{li2022cross}. Notably, the combination of AT + CMT improves the previous best performance (from AT) by +1.0\% mAP on the ``0.02'' split and +2.1\% mAP on the ``All'' split. We observe a relatively larger gain from CMT on the ``All'' split than the ``0.02'' split, and this demonstrates a strong ability to learn robust features from more unlabeled data: In real-world applications, we can obtain abundant unlabeled data but labeling them can be costly. We hope that domain adaptation methods can persistently improve target-domain performance as unlabeled training data grow, and CMT is exactly fitted for this role.

\begin{table}[t]
    \centering
    \begin{tabular}{@{}l|c c@{}}
        \toprule
        Method & AP (Car) & Gain w.r.t. Source \\
        \midrule
        Source$^\dag$ & 40.3 & - \\
        Oracle$^\dag$ & 66.4 & - \\
        \midrule
        MeGA-CDA~\cite{vs2021mega} & 43.0 & +2.7 \\
        TIA~\cite{zhao2022task} & 44.0 & +3.7 \\
        SIGMA~\cite{li2022sigma} & 45.8 & +5.5 \\
        \midrule
        PT~\cite{chen2022learning} & 60.2 & +19.9 \\
        PT~\cite{chen2022learning} + CMT (Ours) & \bf{64.3 (+4.1)} & \bf{+24.0} \\
        \bottomrule
    \end{tabular}\\
    \footnotesize{$^\dag$ Results from PT~\cite{chen2022learning}.}
    \caption{\textbf{Domain adaptation between datasets captured by different cameras (from KITTI to Cityscapes).} Mean-teacher self-training method PT~\cite{chen2022learning} outperforms other methods by a large margin, and our CMT further boosts the target-domain performance by 4.1\% AP. The resulting object detector performs almost as well as a detector directly trained using target-domain labels (``Oracle''). }
    \label{tab:kitti}
    \vspace{-3mm}
\end{table}

\begin{table*}[t]
    \centering
    \vspace{-3mm}
    \setlength{\tabcolsep}{0.7mm}
    \resizebox{\linewidth}{!}{
    \begin{tabular}{@{}l|c c c c c c c c c c c c c c c c c c c c|c@{}}
        \toprule
        Method & aero & bike & bird & boat & bottle & bus & car & cat & chair & cow & table & dog & horse & motor & prsn & plant & sheep & sofa & train & tv & mAP \\
        \midrule
        Source$^\dag$ & 23.0 & 39.6 & 20.1 & 23.6 & 25.7 & 42.6 & 25.2 & 0.9 & 41.2 & 25.6 & 23.7 & 11.2 & 28.2 & 49.5 & 45.2 & 46.9 & 9.1 & 22.3 & 38.9 & 31.5 & 28.8 \\
        Oracle$^\dag$ & 33.3 & 47.6 & 43.1 & 38.0 & 24.5 & 82.0 & 57.4 & 22.9 & 48.4 & 49.2 & 37.9 & 46.4 & 41.1 & 54.0 & 73.7 & 39.5 & 36.7 & 19.1 & 53.2 & 52.9 & 45.0 \\
        \midrule
        ICR-CCR~\cite{xu2020exploring} & 28.7 & 55.3 & 31.8 & 26.0 & 40.1 & 63.6 & 36.6 & 9.4 & 38.7 & 49.3 & 17.6 & 14.1 & 33.3 & 74.3 & 61.3 & 46.3 & 22.3 & 24.3 & 49.1 & 44.3 & 38.3 \\
        HTCN~\cite{chen2020harmonizing} & 33.6 & 58.9 & 34.0 & 23.4 & 45.6 & 57.0 & 39.8 & 12.0 & 39.7  & 51.3 & 20.1  & 20.1& 39.1 & 72.8 & 61.3 & 43.1 & 19.3 & 30.1 & 50.2 & 51.8 & 40.3 \\
        DM~\cite{kim2019diversify} & 25.8 & 63.2 & 24.5 & 42.4 & 47.9 & 43.1 & 37.5 & 9.1 & 47.0 & 46.7 & 26.8 & 24.9 & 48.1 & 78.7 & 63.0 & 45.0 & 21.3 & 36.1 & 52.3 & 53.4 & 41.8 \\
        UMT~\cite{deng2021unbiased} & 39.6 & 59.1 & 32.4 & 35.0 & 45.1 & 61.9 & 48.4 & 7.5  & 46.0 & 67.6 & 21.4 & 29.5 & 48.2 & 75.9 & 70.5 & 56.7 & 25.9 & 28.9 & 39.4 & 43.6 & 44.1 \\
        TIA~\cite{zhao2022task} & 42.2 & 66.0 & 36.9 & 37.3 & 43.7 & 71.8 & 49.7 & 18.2 & 44.9 & 58.9 & 18.2 & 29.1 & 40.7 & 87.8 & 67.4 & 49.7 & 27.4 & 27.8 & 57.1 & 50.6 & 46.3 \\
        \midrule
        AT$^\ddag$~\cite{li2022cross} & 33.1 & 66.1 & 35.3 & 44.9 & 57.5 & 44.9 & 51.0 & 5.8 & 59.5 & 54.9 & 34.6 & 23.5 & 64.3 & 84.0 & 75.4 & 51.5 & 17.1 & 30.3 & 43.3 & 37.2 & 45.7 \\
        AT~\cite{li2022cross} + CMT (Ours) & 39.8 & 56.3 & 38.7 & 39.7 & 60.4 & 35.0 & 56.0 & 7.1 & 60.1 & 60.4 & 35.8 & 28.1 & 67.8 & 84.5 & 80.1 & 55.5 & 20.3 & 32.8 & 42.3 & 38.2 & \bf{47.0 (+1.3)} \\
        \bottomrule
    \end{tabular}}\\
    \footnotesize{$^\dag$ Results from AT~\cite{li2022cross}. $^\ddag$ Results reproduced using the released code by AT~\cite{li2022cross}.}
    \caption{\textbf{Domain adaptation from realistic images (Pascal VOC) to artistic images (Clipart1k).} Our CMT improves upon AT~\cite{li2022cross} and achieves the new best overall accuracy of 47.0\% mAP.}
    \label{tab:clipart}
    \vspace{-3mm}
\end{table*}

\subsection{Across Cameras}
Real-world sensors like cameras have drastically different configurations (\eg, intrinsic parameters, resolutions), and such differences can adversely affect the deployed object detectors. In addition, Cityscapes is collected from multiple cities different from KITTI, so the street scenes exhibit more diversity and bring more challenge to this task. We evaluate CMT on the KITTI $\rightarrow$ Cityscapes domain adaptation benchmark to study its effectiveness in cross-camera adaptation. Following the practice of previous work, we only train and evaluate object detectors for the common category ``Car'' shared by KITTI and Cityscapes. The results are compared in Table~\ref{tab:kitti}. The mean-teacher self-training method PT outperforms all previous methods by a large margin (about 15\% AP). Moreover, when combined with our proposed CMT framework, PT receives an additional 4.1\% AP performance improvement.

\subsection{Realistic to Artistic}
We also study a domain adaptation task with different image styles, from realistic images to artistic images. Here we use Pascal VOC as the source domain dataset, which contains images captured in natural scenes. The object detector is adapted to the target domain of Clipart, an artistic image dataset, without any human supervision. Table~\ref{tab:clipart} shows the results in this domain adaptation task. The combination of AT + CMT improves AT by 1.3\% mAP, and outperforms the previous best TIA~\cite{zhao2022task} by 0.7\% mAP.

\subsection{Analysis and Ablation Study}
In this section, we provide additional experimental results to understand the source of performance gain in our proposed approach CMT. We use the challenging Cityscapes $\rightarrow$ Foggy Cityscapes benchmark (``All'' split) as an example, and conduct experiments with AT~\cite{li2022cross} and PT~\cite{chen2022learning} base methods.

\noindent\textbf{Noise in pseudo-labels.} One benefit of contrastive learning is that it does not require accurate class labels. By discriminating each individual instance~\cite{chen2020simple,he2020momentum} or clusters of instances~\cite{caron2020unsupervised}, contrastive learning optimizes their visual representations. Two facts are worth noting in contrastive learning: 1) A pair of instances from the same class may form a negative pair and thus are pushed apart. 2) A cluster of instances found by learned features may not coincide with an actual class defined by humans, and instances in it are still pulled together. Yet, contrastive learning can still acquire robust and reliable visual representation for down-stream tasks. This observation suggests that contrastive learning is tolerant to noisy pseudo-labels.

\begin{figure}
    \centering
    \vspace{-3mm}
    \includegraphics[width=\columnwidth]{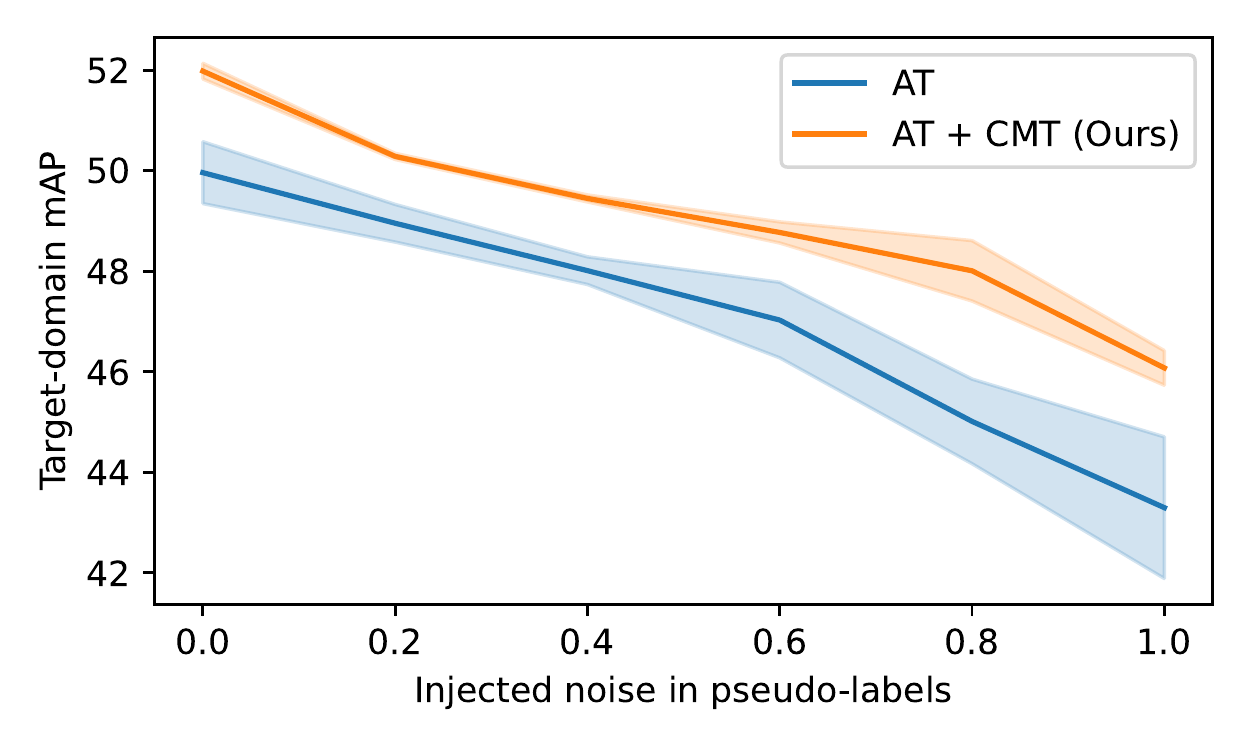}
    \caption{\textbf{Impact of pseudo-label noise on Foggy Cityscapes target-domain performance.} Shadows depict the standard deviation across three runs. Baseline AT suffers from injected noise (implemented by randomly perturbing class labels). CMT helps recover the accuracy from noisy pseudo-labels and reduce performance instability.}
    \label{fig:noise}
    \vspace{-3mm}
\end{figure}

To demonstrate our object-level contrastive learning in CMT can extract beneficial learning signals from pseudo-labels even if they are noisy, we design the following analytical experiment: In each training iteration of AT, we manually perturb the pseudo-labels generated by the teacher before using them for the contrastive loss and detection loss. Specifically, for a fraction (ranging from 20\% to 100\%) of the predicted objects, we re-assign a random class label to them. Thus, the quality of pseudo-labels is affected by the injected noise and will harm the domain adaptation pipeline.

The results of this analytical experiment are shown in Figure~\ref{fig:noise}. As we inject more noise into the pseudo-labels, the target-domain performance of AT drops considerably. The accuracy does not decrease to a random-prediction level, because the model still receives correct supervision from source-domain labels. By contrast, CMT utilizes object-level contrastive learning to combat the pseudo-label noise and partially recover the target-domain performance from two aspects: First, CMT reduces performance variance across multiple runs, resulting in greater stability in the presence of noisy pseudo-labels. Specifically, CMT reduces the standard deviation of performance from 1.4\% to 0.4\% when the level of noise is at 1.0. Second, as the level of pseudo-label noise increases, CMT provides larger performance gains. For example, when the injected noise increases from 0.0 to 1.0, the mean performance gain increases from +2.0\% to +2.8\%. This phenomenon demonstrates that our object-level contrastive learning is able to exploit helpful information from pseudo-labels with noise for unsupervised domain adaptation.

\begin{table}[t]
    \centering
    \resizebox{\linewidth}{!}{
    \begin{tabular}{@{}l|c c|c c@{}}
        \toprule
        \multirow{2}{*}{Method} & Class-based & Multi-scale & \multirow{2}{*}{mAP} & Gain \\
        & Contrast & Features & & w.r.t. PT \\
        \midrule
        PT & - & - & 47.1 & - \\
        \midrule
        & \ding{55} & \ding{55} & 47.8 & +0.7 \\
        PT + CMT & \ding{55} & \ding{51} & 48.2 & +1.1 \\
        (Ours) & \ding{51} & \ding{55} & 48.7 & +1.6 \\
        & \ding{51} & \ding{51} & \bf{49.3} & \bf{+2.2} \\
        \bottomrule
    \end{tabular}}\\
    \caption{\textbf{Ablation study of components in object-level contrastive learning.} Our proposed CMT improves the performance of PT in the Foggy Cityscapes target domain. There are two key designs in our object-level contrastive learning: 1) contrasting object-level features based on the predicted classes in pseudo-labels (Equation~\ref{eq:contrast}), and 2) learning multi-scale features from various backbone stages. Class-based contrast brings more performance gain as compared with multi-scale features, and their combination leads to a further improvement.}
    \label{tab:ablation}
    \vspace{-3mm}
\end{table}

\noindent\textbf{Components in object-level contrastive learning.} As described in Section~\ref{sec:obj-contrast}, our object-level contrastive learning has two design choices for exploiting the pseudo-labels: 1) class-based contrast, and 2) multi-scale features. Here, we dissect these components and use the example of PT + CMT on Cityscapes $\rightarrow$ Foggy Cityscapes to observe the performance gain from each component.

We summarize the results in Table~\ref{tab:ablation}. The vanilla object-level contrastive learning without the two additional designs is already helpful to PT, demonstrating its effectiveness in feature adaptation. Class-based contrast brings more performance gain than learning multi-scale features (+1.6\% vs. +1.1\% mAP). Furthermore, when the two designs function jointly, an additional performance gain is achieved.

\begin{figure}
    \centering
    \includegraphics[width=\columnwidth]{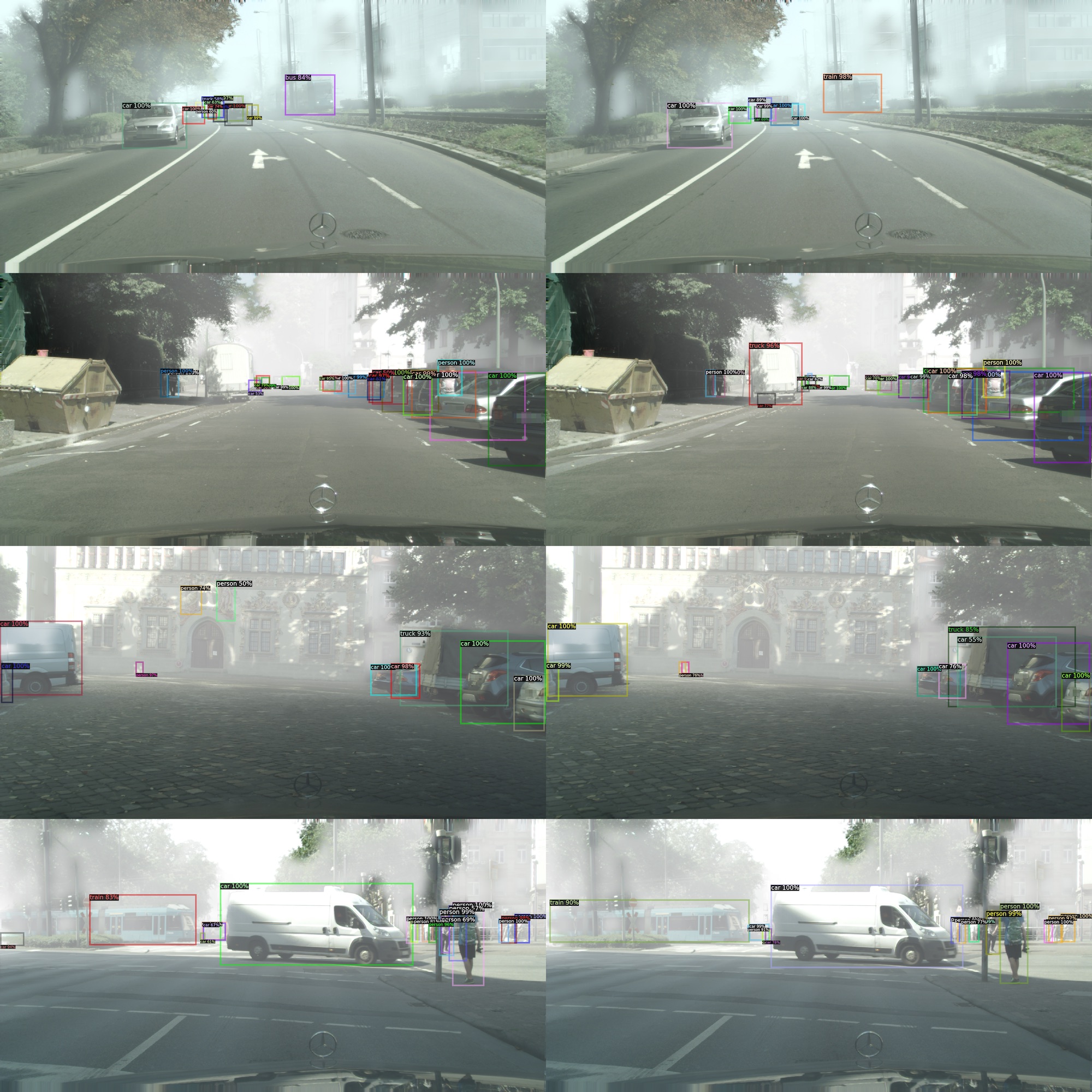}
    \caption{\textbf{Qualitative results.} We compare the detection results of AT (\textbf{left}) and AT + CMT (\textbf{right}) on Foggy Cityscapes. CMT fixes errors of mis-classification (row 1, the train), false negative (row 2, the truck), and false positive (row 3, the person-like sculptures), and improves the localization (row 4, the train).}
    \label{fig:qual}
    \vspace{-3mm}
\end{figure}

\noindent\textbf{Qualitative results.} Finally, we provide some detection visualizations to intuitively demonstrate the benefit of CMT. We compare AT and AT + CMT on the challenging Cityscapes $\rightarrow$ Foggy Cityscapes benchmark. As shown in Figure~\ref{fig:qual}, the better object-level representations learned by CMT assist the detector to distinguish foreground object categories and better locate them. More high-resolution visualization is presented in the supplementary material.

\section{Conclusion}
\label{sec:conclusion}
In this work, we identify the intrinsic alignment between contrastive learning and mean-teacher self-training, and propose \emph{Contrastive Mean Teacher}, an integrated unsupervised domain adaptation framework. Extensive experiments show that our object-level contrastive learning consistently improves several existing methods and achieves state-of-the-art results on multiple benchmarks.
There are several interesting future directions: 1) developing unsupervised domain adaptation methods for more challenging real-world data with diverse types of domain shifts, 2) selecting or prioritizing objects in object-level contrastive learning according to their significance, and 3) integrating contrastive learning with source-free domain adaptation~\cite{li2021free,li2022source}.

\noindent{\footnotesize\textbf{Acknowledgement.} This work was supported in part by the IBM-Illinois Discovery Accelerator Institute, NSF Grant 2106825, NIFA Award 2020-67021-32799, and the NCSA Fellows program. This work used NVIDIA GPUs at NCSA Delta through allocation CIS220014 from the ACCESS program.  We appreciate the helpful discussion with Yu-Jhe Li. \par}

\clearpage

{\small
\bibliographystyle{ieee_fullname}
\bibliography{egbib}
}

\clearpage
\onecolumn
\appendix
\label{sec:app}
\section{Additional Visualization Results}
\begin{figure}[!h]
    \centering
    \begin{subfigure}[b]{0.3\columnwidth}
        \centering
        \includegraphics[height=6cm]{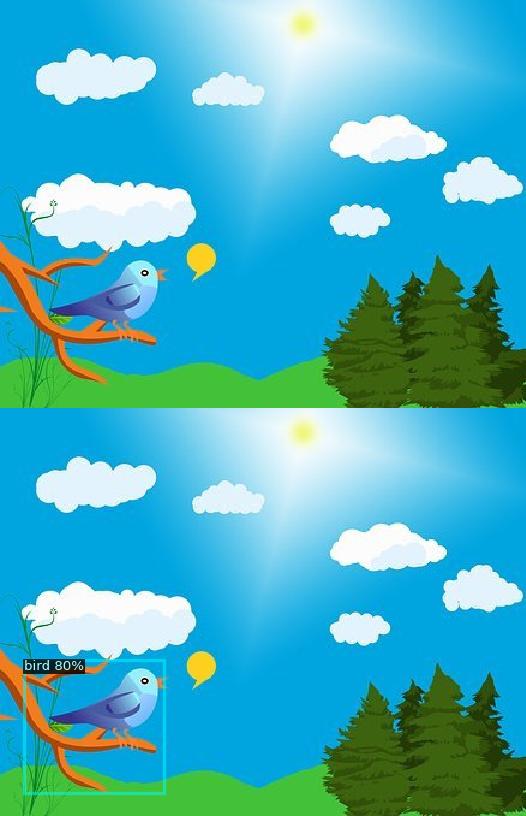}
        \caption{CMT detects the bird missed by the baseline.}
    \end{subfigure}
    \quad
    \begin{subfigure}[b]{0.3\columnwidth}
        \centering
        \includegraphics[height=6cm]{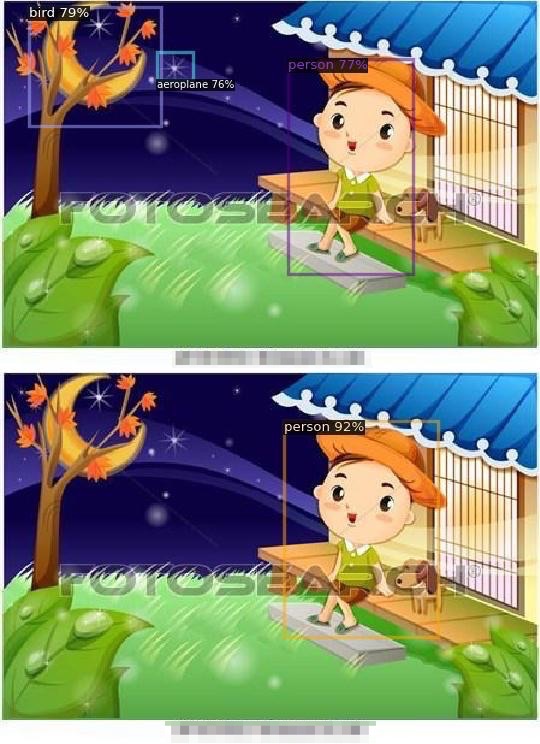}
        \caption{CMT avoids detecting the moon and star as bird and aeroplane.}
    \end{subfigure}
    \quad
    \begin{subfigure}[b]{0.3\columnwidth}
        \centering
        \includegraphics[height=6cm]{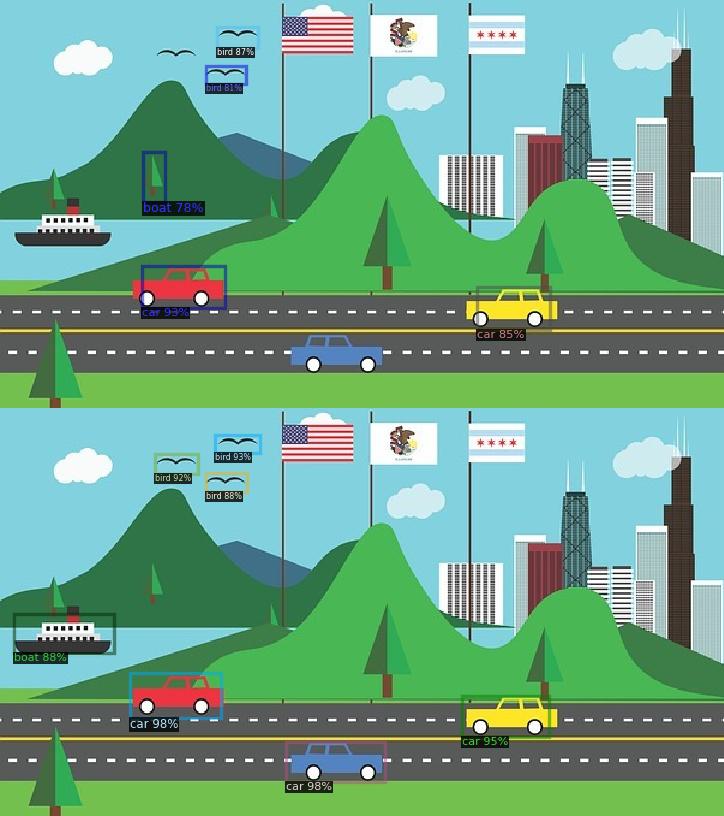}
        \caption{CMT detects the boat and birds missed by the baseline.}
    \end{subfigure}
    
    \begin{subfigure}[b]{0.3\columnwidth}
        \centering
        \includegraphics[height=6cm]{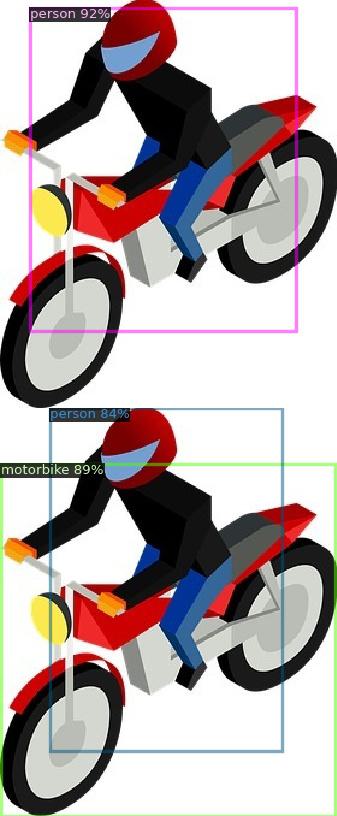}
        \caption{CMT detects the motorbike missed by the baseline.}
    \end{subfigure}
    \quad
    \begin{subfigure}[b]{0.3\columnwidth}
        \centering
        \includegraphics[height=6cm]{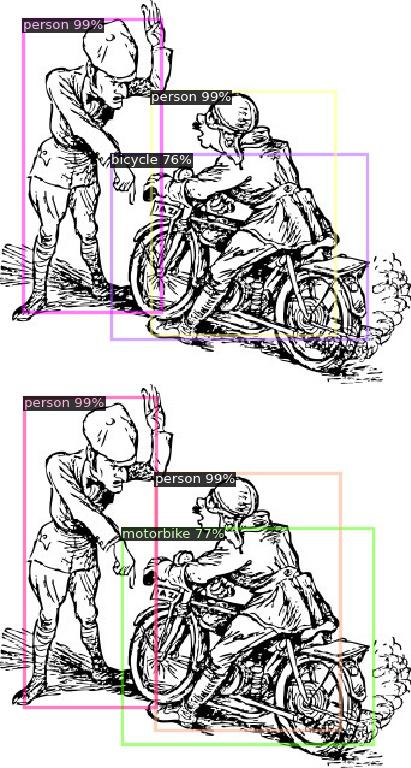}
        \caption{CMT fixes the mis-classification of the motorbike.}
    \end{subfigure}
    \quad
    \begin{subfigure}[b]{0.3\columnwidth}
        \centering
        \includegraphics[height=6cm]{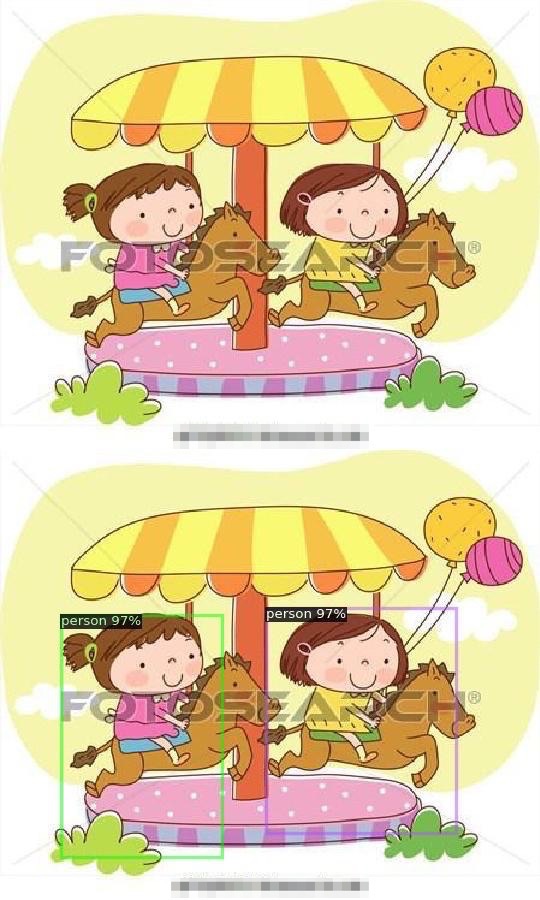}
        \caption{CMT detects the persons missed by the baseline.}
    \end{subfigure}
    
    \caption{\textbf{Additional qualitative results from Clipart1k.} In the visualized images, AT (\textbf{top} in each sub-figure) makes some incorrect predictions, while AT + CMT (\textbf{bottom} in each sub-figure) can correct them.}
    \label{fig:supp-clipart}
\end{figure}

\begin{figure}
    \centering
    \vspace{-8mm}
    \begin{subfigure}[b]{0.4\columnwidth}
        \centering
        \includegraphics[width=\columnwidth]{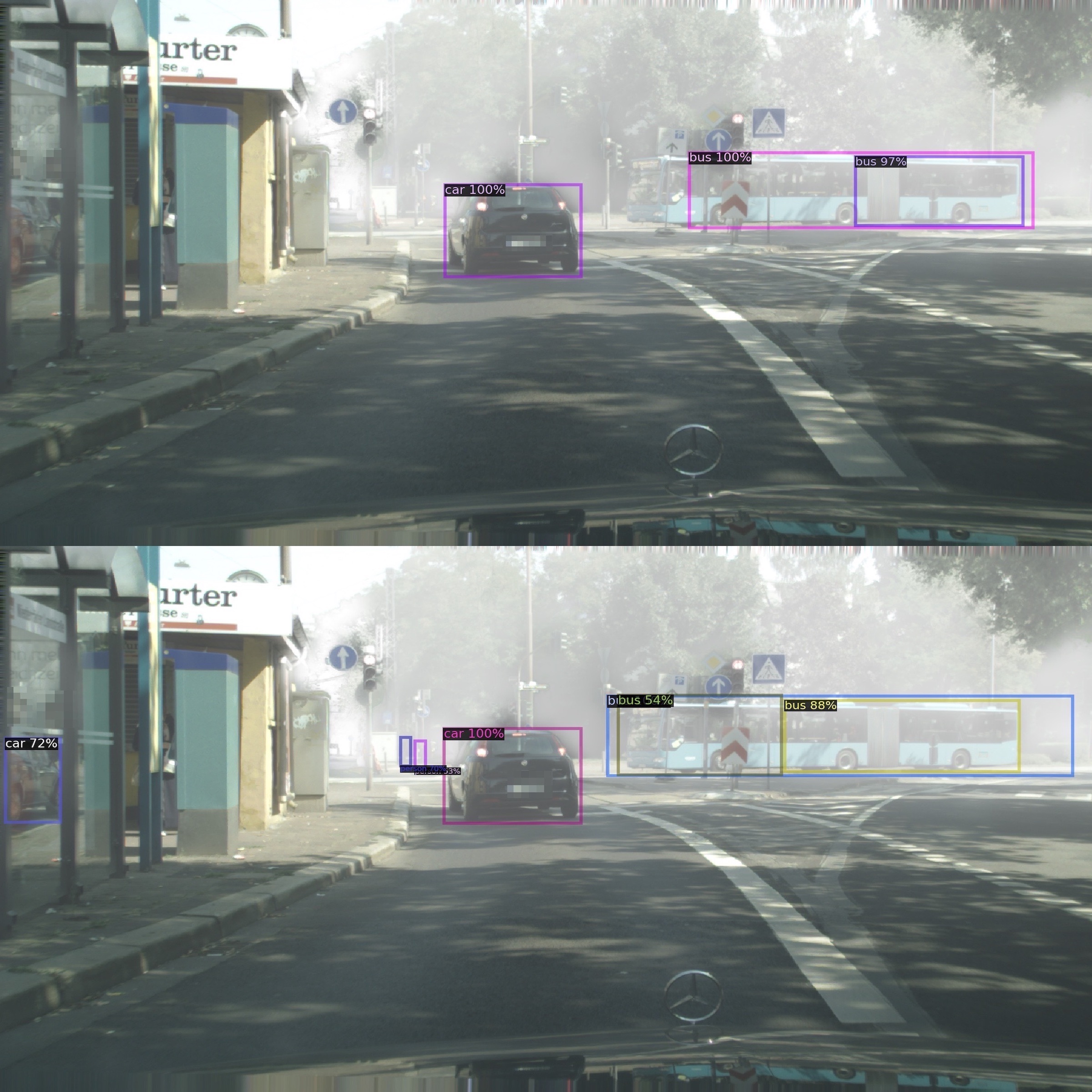}
        \caption{CMT detects the car in the mirror and distant persons.}
    \end{subfigure}
    \quad
    \begin{subfigure}[b]{0.4\columnwidth}
        \centering
        \includegraphics[width=\columnwidth]{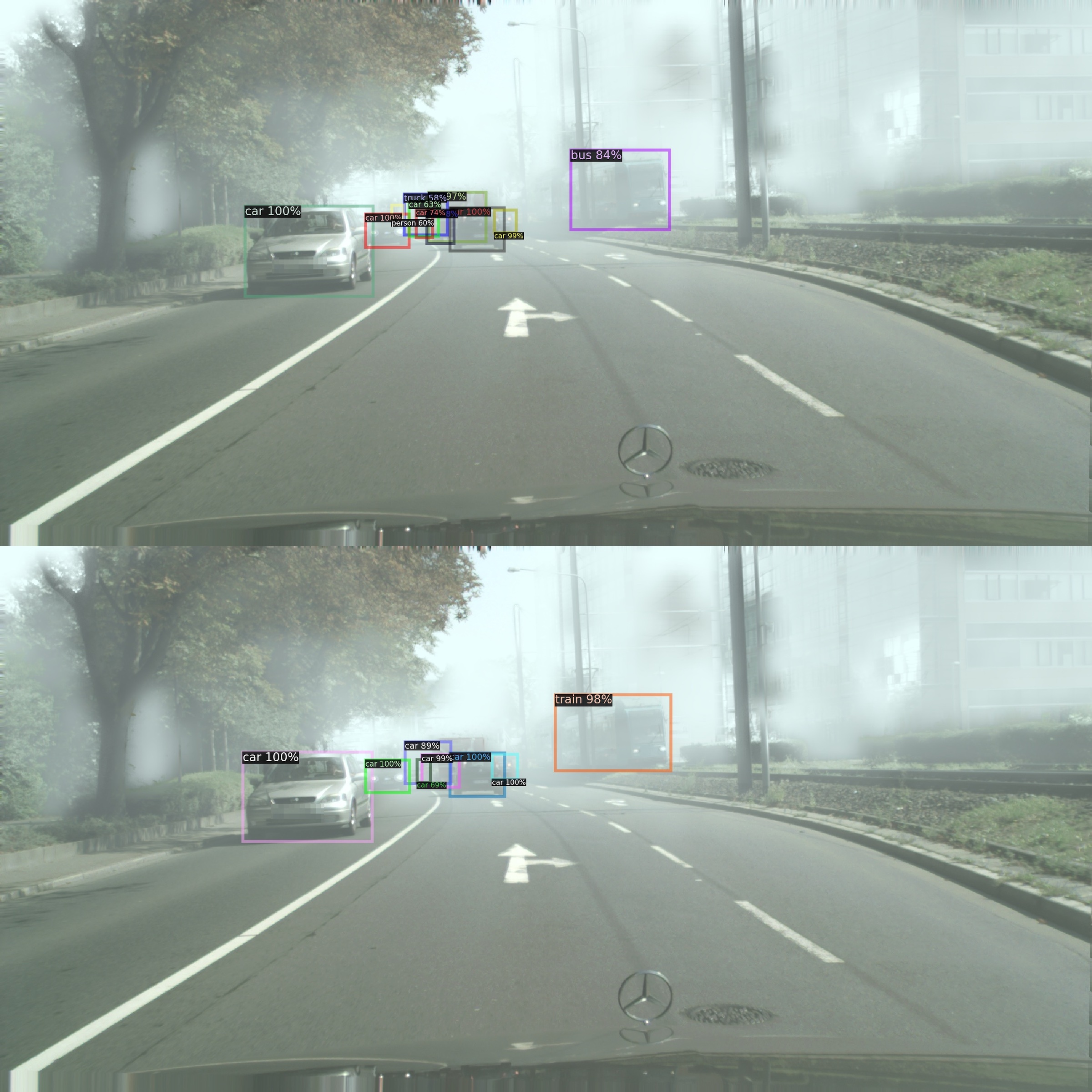}
        \caption{CMT fixes the mis-classification of the train.}
    \end{subfigure}
    
    \begin{subfigure}[b]{0.4\columnwidth}
        \centering
        \includegraphics[width=\columnwidth]{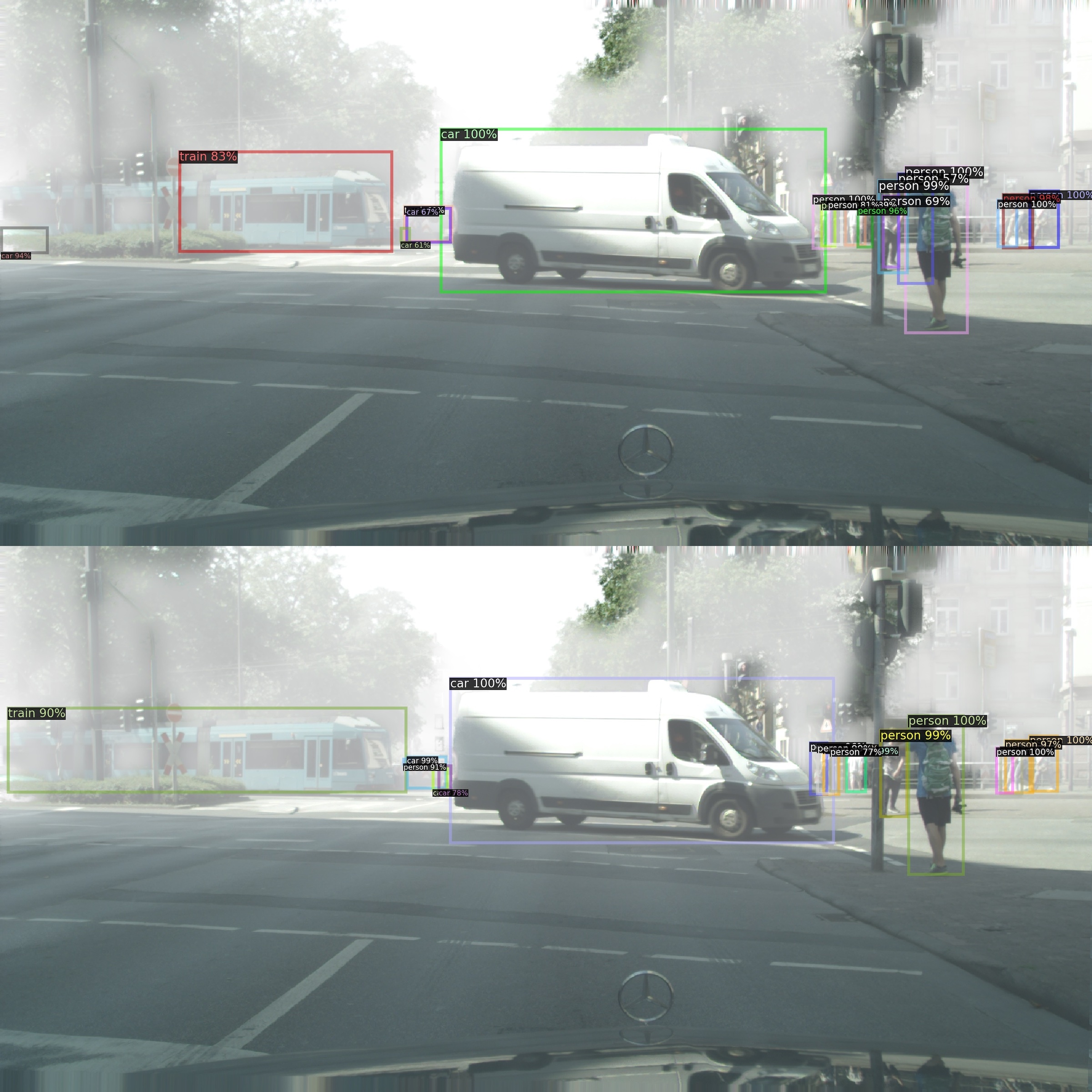}
         \caption{CMT provides a more precise bounding box of the train.}
    \end{subfigure}
    \quad
    \begin{subfigure}[b]{0.4\columnwidth}
        \centering
        \includegraphics[width=\columnwidth]{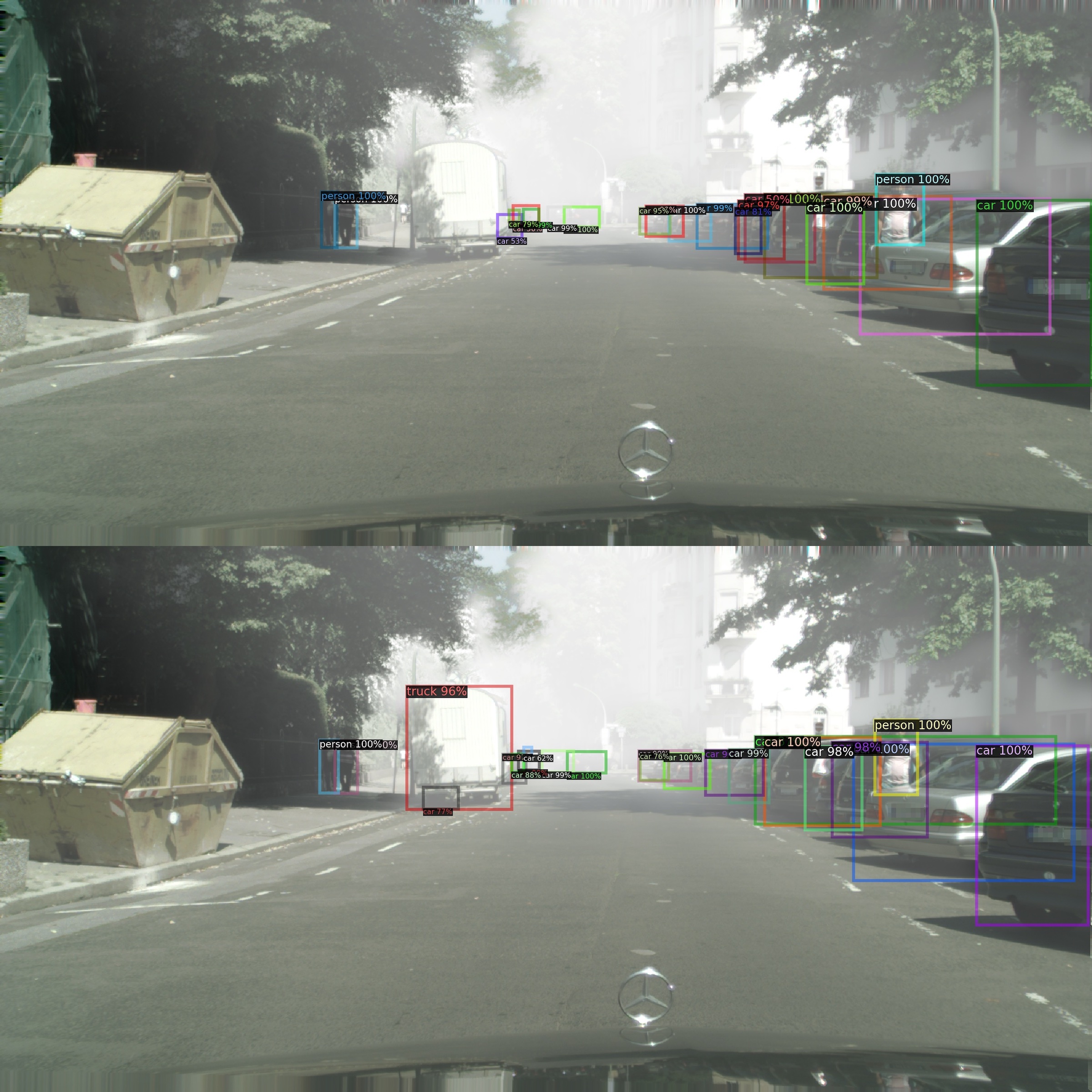}
        \caption{CMT detects the truck missed by the baseline.}
    \end{subfigure}
    
    \begin{subfigure}[b]{0.4\columnwidth}
        \centering
        \includegraphics[width=\columnwidth]{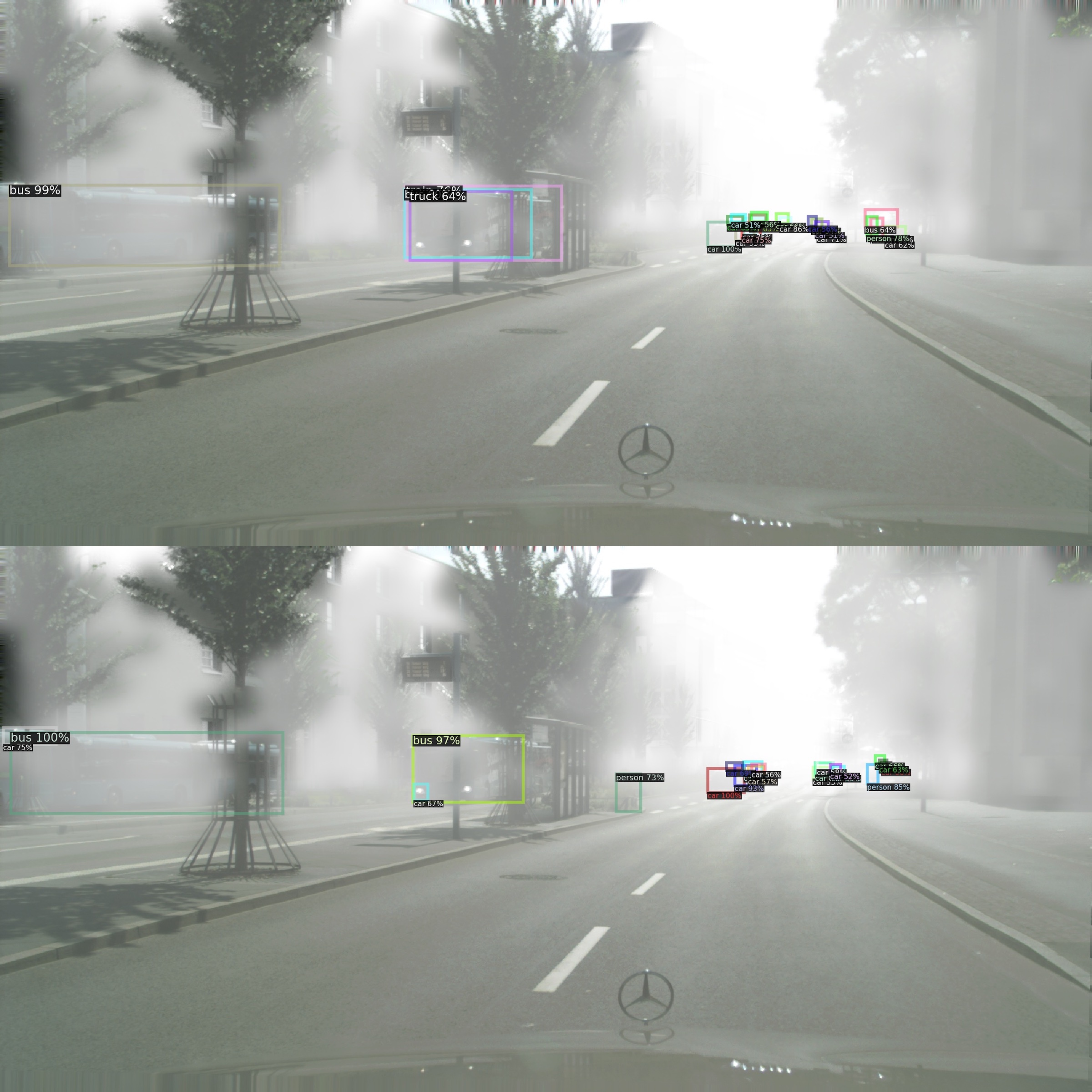}
        \caption{CMT fixes the mis-classification of the bus.}
    \end{subfigure}
    \quad
    \begin{subfigure}[b]{0.4\columnwidth}
        \centering
        \includegraphics[width=\columnwidth]{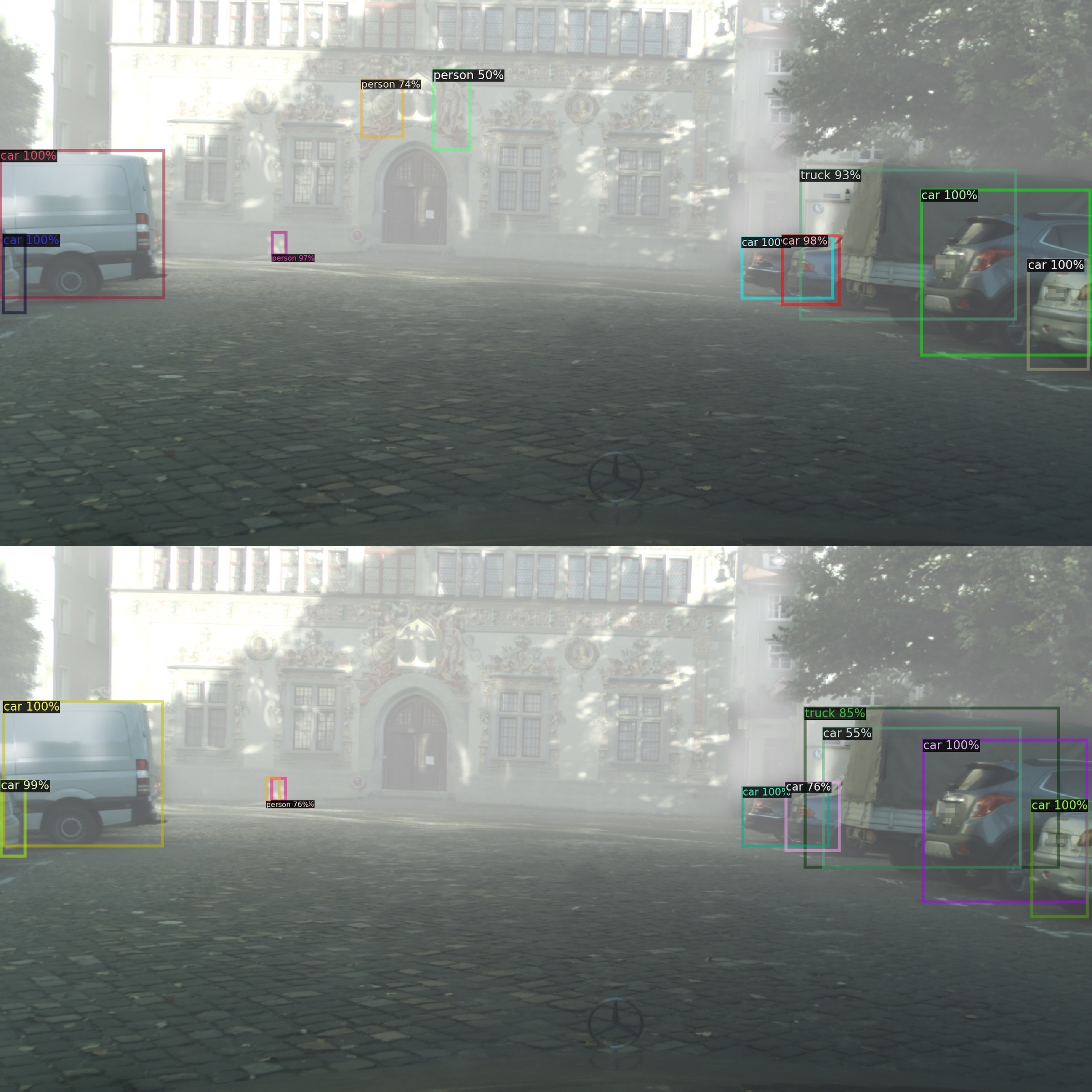}
        \caption{CMT avoids detecting the sculptures as real persons.}
    \end{subfigure}
    \caption{\textbf{Additional qualitative results from Foggy Cityscapes.} In the visualized images, AT (\textbf{top} in each sub-figure) makes some incorrect predictions, while AT + CMT (\textbf{bottom} in each sub-figure) can correct them.}
    \label{fig:supp-foggy}
\end{figure}

We present more high-resolution visualization to compare the baseline Adaptive Teacher (AT)~\cite{li2022cross} and our AT + CMT qualitatively on the Pascal VOC $\rightarrow$ Clipart1k benchmark in Figure~\ref{fig:supp-clipart}, and on the Cityscapes $\rightarrow$ Foggy Cityscapes benchmark in Figure~\ref{fig:supp-foggy}. Each pair of images show results by AT (\textbf{top}) and AT + CMT (\textbf{bottom}).

\clearpage

\section{Pseudo-code for Contrastive Mean Teacher}
In Algorithm~\ref{alg:cmt}, we present a pseudo-code outline of our Contrastive Mean Teacher (CMT) framework. The key difference between CMT and a traditional Mean Teacher (MT)~\cite{tarvainen2017mean} framework is \textcolor{teal}{highlighted}.

\vspace{-3mm}
\begin{algorithm}[h!]
    \SetAlgoLined
    \SetInd{0.1em}{2em}
    \KwInput{Object detectors: Student $D(\cdot;\theta^\mathcal{Q})$ and Teacher $D(\cdot;\theta^\mathcal{K})$. For consistency with prior work, we use letter $\mathcal{Q}$ for student-related variables, and $\mathcal{K}$ for teacher-related ones. We denote their feature extraction modules as $f(\cdot;\theta^\mathcal{Q})$ and $f(\cdot;\theta^\mathcal{K})$. Hyper-parameters: Momentum $\alpha$ in exponential moving average (EMA), pseudo-label score threshold $\gamma$, temperature $\tau$ in contrastive loss, loss weights $\lambda_\text{contrast},\lambda_\text{unsup\_det},\lambda_\text{sup\_det}$, and learning rate $\eta$.}
    \KwOutput{Student $D(\cdot;\theta^\mathcal{Q})$ and Teacher $D(\cdot;\theta^\mathcal{K})$ after unsupervised domain adaptation.}
    \For{$iteration\gets 1$ \KwTo $T_\text{max\_iterations}$}
    {
        \tcp{1. Load data mini-batch}
        Get batch of source-domain labeled images $\mathcal{I}^\text{labeled}$, corresponding bounding boxes $\mathcal{B}^\text{labeled}$, and classes $\mathcal{C}^\text{labeled}$

        Get batch of target-domain unlabeled images $\mathcal{I}^\text{unlabeled}$

        Student's strong augmentation: $\mathcal{I}^{\text{labeled},\mathcal{Q}}=t^\mathcal{Q}(\mathcal{I}^\text{labeled}), \mathcal{I}^{\text{unlabeled},\mathcal{Q}}=t^\mathcal{Q}(\mathcal{I}^\text{unlabeled})$

        Teacher's weak augmentation: $\mathcal{I}^{\text{labeled},\mathcal{K}}=t^\mathcal{K}(\mathcal{I}^\text{labeled}), \mathcal{I}^{\text{unlabeled},\mathcal{K}}=t^\mathcal{K}(\mathcal{I}^\text{unlabeled})$

        \tcp{2. Update Teacher}
        Update Teacher by EMA: $\theta^\mathcal{K}=\alpha\theta^\mathcal{K} + (1-\alpha)\theta^\mathcal{Q}$

        \tcp{3. Pseudo-label}
        Generate pseudo-labels with Teacher detector: $\mathcal{B}^\text{unlabeled},\mathcal{C}^\text{unlabeled}=\text{Filter}\left(D(\mathcal{I}^{\text{unlabeled},\mathcal{K}}; \theta^\mathcal{K}), \gamma \right)$

        {\color{teal}
        \tcp{4. Compute multi-scale feature maps}

        \tcp{In practical implementation, feature maps are obtained from forward passes needed for pseudo-labels and unsupervised detection loss, so there is no computation overhead.}
        Compute Student's features: $\mathcal{F}^\mathcal{Q}=f(\mathcal{I}^{\text{unlabeled},\mathcal{Q}};\theta^\mathcal{Q})$

        Compute Teacher's features: $\mathcal{F}^\mathcal{K}=f(\mathcal{I}^{\text{unlabeled},\mathcal{K}};\theta^\mathcal{K})$

        \tcp{5. Unsupervised branch: object-level contrastive loss}
        (Optional) Post-processing pseudo-labels: $\mathcal{B}^\text{unlabeled},\mathcal{C}^\text{unlabeled}=\text{PostProc}(\mathcal{B}^\text{unlabeled},\mathcal{C}^\text{unlabeled})$

        Get number of objects: $N=\text{len}(\mathcal{B}^\text{unlabeled})$

        \tcp{Each level of multi-scale features}
        \For{$k\gets 1$ \KwTo $K_\text{max\_levels}$}
        {
            \tcp{Each object}
            \For{$i\gets 1$ \KwTo $N$}
            {
                Locate Student's object-level features: $z^\mathcal{Q}_{k,i}=\text{Normalize}(\text{ROIAlign}(F_k^\mathcal{Q},B_i^\text{unlabeled}))$

                Locate Teacher's object-level features: $z^\mathcal{K}_{k,i}=\text{Normalize}(\text{ROIAlign}(F_k^\mathcal{K},B_i^\text{unlabeled}))$
            }
            Compute contrastive loss according to Equation~\ref{eq:contrast}: $L_{\text{contrast},k}=\mathcal{L}_\text{contrast}\left(\{z^\mathcal{Q}_{k,1},\dots,z^\mathcal{Q}_{k,N}\},\{z^\mathcal{K}_{k,1},\dots,z^\mathcal{K}_{k,N}\},\{C_1^\text{unlabeled},\dots,C_N^\text{unlabeled}\},\tau\right)$
        }
        Compute total contrastive loss: $L_{\text{contrast}}=\sum_{k=1}^{K_\text{max\_levels}}L_{\text{contrast},k}$
        }

        \tcp{6. Unsupervised branch: detection loss}
        Compute unsupervised detection loss: $L_\text{unsup\_det}=\mathcal{L}_\text{det}(D(\mathcal{I}^{\text{unlabeled},\mathcal{Q}};\theta^\mathcal{Q}), \mathcal{B}^\text{unlabeled},\mathcal{C}^\text{unlabeled})$

        \tcp{7. Supervised branch: detection loss}
        Compute supervised detection loss: $L_\text{sup\_det}=\mathcal{L}_\text{det}(D(\mathcal{I}^{\text{labeled},\mathcal{Q}};\theta^\mathcal{Q}), \mathcal{B}^\text{labeled},\mathcal{C}^\text{labeled})$

        \tcp{8. Optimize}
        Compute total loss: $L=\lambda_\text{contrast}L_\text{contrast}+\lambda_\text{unsup\_det}L_\text{unsup\_det}+\lambda_\text{sup\_det}L_\text{sup\_det}$

        Take SGD step: $\theta^\mathcal{Q}=\theta^\mathcal{Q}-\eta\nabla_{\theta^\mathcal{Q}} L$
    }
    \caption{Contrastive Mean Teacher}
    \label{alg:cmt}
\end{algorithm}
\vspace{-5mm}

\end{document}